# AI based signage classification for linguistic landscape studies

**Authors:** Yuqin Jiang[*], Song Jiang, Jacob Algrim[†], Trevor Harms[†], Maxwell Koenen[†], Xinya Lan[†], Xingyu Li[†], Chun-Han Lin[†], Jia Liu[†], Jiayang Sun[†], Henry Zenger[†]

University of Hawaiʻi at Mānoa

[*] Corresponding author: yuqinj@hawaii.edu
[†] Authors with equal contributions.

## Abstract

Linguistic Landscape (LL) research traditionally relies on manual photography and annotation of public signages to examine distribution of languages in urban space. While such methods yield valuable findings, the process is time-consuming and difficult for large study areas. This study explores the use of AI powered language detection method to automate LL analysis. Using Honolulu Chinatown as a case study, we constructed a georeferenced photo dataset of 1,449 images collected by researchers and applied AI for optical character recognition (OCR) and language classification. We also conducted manual validations for accuracy checking. This model achieved an overall accuracy of 79%. Five recurring types of mislabeling were identified, including distortion, reflection, degraded surface, graffiti, and hallucination. The analysis also reveals that the AI model treats all regions of an image equally, detecting peripheral or background texts that human interpreters typically ignore. Despite these limitations, the results demonstrate the potential of integrating AI-assisted workflows into LL research to reduce such time-consuming processes. However, due to all the limitations and mis-labels, we recognize that AI cannot be fully trusted during this process. This paper encourages a hybrid approach combining AI automation with human validation for a more reliable and efficient workflow.

## 1. Introduction

Linguistic landscape refers to the visibility and salience of languages in public space. As Ben-Rafael et al. (2006) noted, the linguistic landscape "constitutes the very scene made of streets, corners, circuses, parks, buildings where society's public life takes place. As such, this scene carries crucial socio-symbolic importance as it actually identifies and thus serves as the emblem of societies, communities, and regions." Public signages, be it in the form of advertisements, storefronts, or building names, carry crucial sociolinguistic significance, as they both reflect and shape the demographic, language, and cultural diversity of the area.

Despite their values, the systematic analysis of signages for linguistic landscapes remains a labor-intensive task. Traditional approaches often rely on manual photography and annotation, where researchers must not only capture images in the field, but also label and categorize signs by language, content, and functions. Partially due to the relative nascence of the field, there still lacks a unified approach to treating data categorization in the linguistic landscape, and data analysis remains highly customized to the specific studies. The categorization process, or sometimes termed coding process, becomes especially challenging in multilingual settings, where signs may include multiple scripts, stylized fonts, or handwritten elements that complicate language identification. Moreover, variations in lighting, sign conditions, and layout further hinder consistent analysis at a large scale.

The site of this study Honolulu Chinatown, an ethnic neighborhood which has played important roles both economically and culturally for the city of Honolulu and the broader Oahu Island where it is located. Ethnic enclaves, and Chinatowns as a genre of their own, have become a major vein of research within the LL field (e.g., Amos, 2021; Leeman & Modan, 2009; Lou, 2010, 2012; Jazul & Bernardo, 2017; Zhang, 2020). Honolulu Chinatown stands out as unique in a number of ways, such as being the only island Chinatown in the United States, the Chinese population was established in Hawai'i well before it was under United States' governance, and in terms of semiotics it lacks the iconic Chinatown arch which marks the boundary of many Chinatowns around the world. A majority of the early Chinese immigrants which led to the established success of the neighborhood initially came as plantation laborers between 1852 and 1899 from southern China. In 1884, 80% of Chinese in Hawai'i worked in rural labor, but this dropped to 40% by 1920 as many Chinese finished their plantation contracts and moved into cities, if they chose not to return to China (Riley, 2024).

While the presence of Chinese languages (including both traditional and simplified character forms, as well as romanizations) and other cultural semiotic materials are quite visible throughout Honolulu Chinatown, it is quite easily classifiable as what Blommaert (2013) calls "superdiverse," that is that it is rich with cultural and linguistic diversity (and sometime economic disparity), such as the north end of Chinatown at times appearing to be more of a Vietnamese neighborhood. As is typical of such superdiverse neighborhoods, you even see different time functions; in the mornings the streets are full of shoppers purchasing fresh local produce and various imported Asian goods, while a shift occurs in the late afternoon as the ethnic groceries close for the day and the neighborhood becomes more of a nightlife district in the evenings. Such a neighborhood offers a rich and complex set of signage and semiotic resources for the present study.

The objectives of this study include (1) to build a georeferenced photo database for LL studies and (2) validate the accuracy of using AI-powered language detection model for LL signage automatically detection.

## 2. Data and methods
### 2.1. Data collection
The photographs were taken by students through a graduate-level seminar. A total of 9 students walked through the neighborhoods in Honolulu Chinatown and captured images of signages that reflected the area's cultural and linguistic landscapes. The goal of this is not only for AI modeling training, but also to create a public image database for the linguist landscape research community.

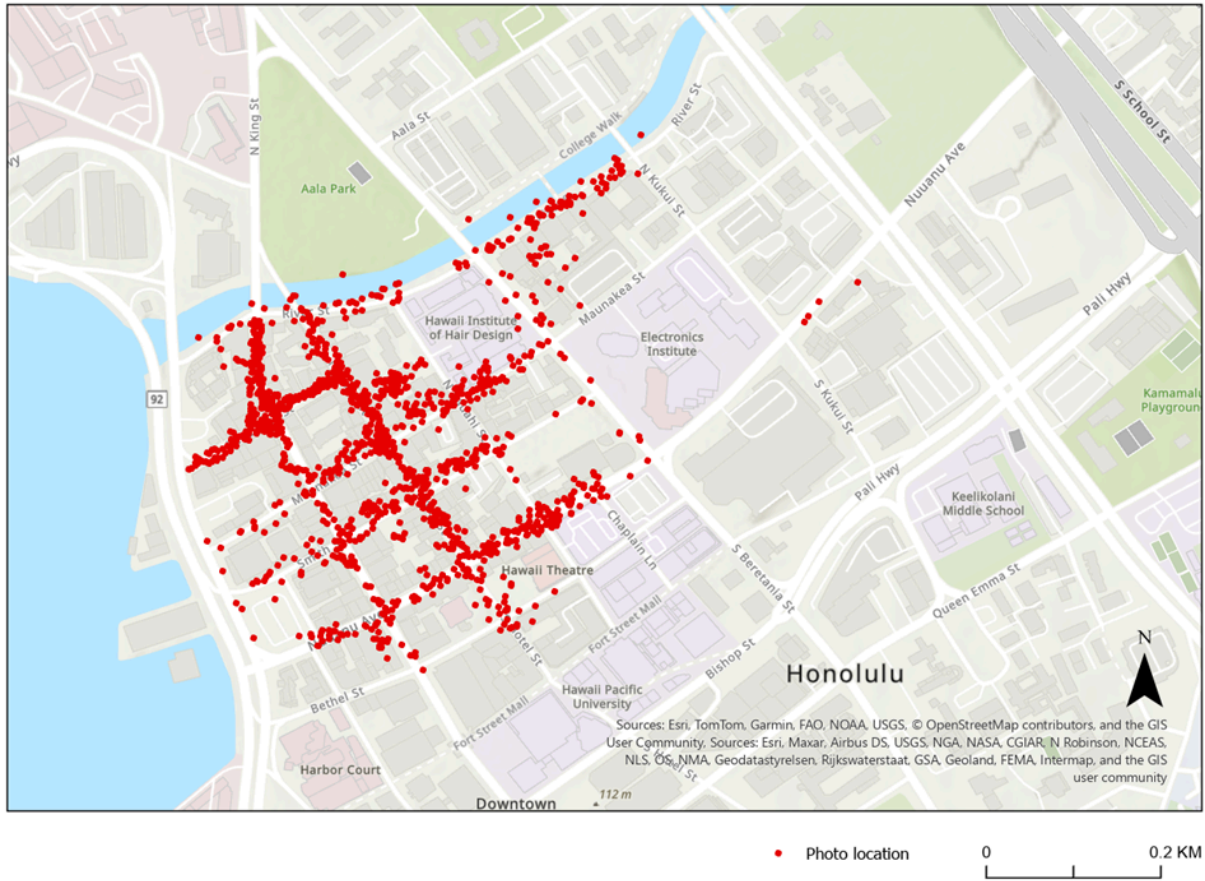

Figure 1. A map of Honolulu Chinatown.

Honolulu Chinatown is located on the western edge of downtown Honolulu. It is generally considered to be bordered by Nuuanu Stream to the west and Bethel Street to the east, between Honolulu Harbor (south) and Beretania Street (north). "The first newspaper reference of the name 'Chinatown' occurs in 1876 (Coover, 2022), and it was called 'Old Chinatown' by 1909" (Coover, 2022, p.II). Maunakea Street is known for its historic business such as lei shops, and food business such as restaurants, markets, grocery stores (Godvin, 2007).

One group was responsible for Maunakea Marketplace, Kekaulike Plaza, Kekaulike street, sections along Nimitz Highway, River Street and the north end of King Street. Maunakea Marketplace and Kekaulike Plaza are both hubs of morning activity in Chinatown. Kekaulike Plaza is a short walking street that is host to shops on either side, and a wet-market-style area that sells fresh produce. Maunakea Marketplace, similarly houses a wet market where produce, fish, and meat is available, but also an indoor food court with Chinese, Filipino, and Nepalese food stalls. There is an open courtyard with public seating in the center of Maunakea Marketplace that hosts two milk tea shops, a Vietnamese hair salon, and various shops selling trinkets, Hawaiian and Chinese-themed goods. Both the Plaza and the Marketplace have attached residential sections, thus potentially providing regular consumers locally along with those who come from other neighborhoods outside of Chinatown to shop.

2.2.     Data organizing and language detection

After collection, all the photographs were organized into a structured dataset to ensure consistency in the subsequent analysis. Each photo's metadata was extracted, which includes date, time, photographer, and location. Additional information, such as the placement of signs (e.g., store front) and sign type (e.g., official or unofficial), was also added into the database as a separated field. While this study only focuses on language detection, as it is the most essential and commonly coded feature of the linguistic landscape, the structure of this database is organized in a way that allows more flexibility for adding more information for various analyses.

Two methods were used to capture the geographical coordinates of each photograph, depending on student preference and device setup. Some students enabled the GPS location detection when taking photos using their smartphones, which automatically embedded latitude and longitude information into the photo's metadata. Other students chose to use EpiCollect5, a mobile application available on both iOS and Android platforms, which allows users to take photos and record spatial information. Photos collected with EpiCollect5 were uploaded to a cloud server along with the associated coordinates. We used the raw coordinate of each photo to ensure accurate map displays. However, cellphone GPS drafting cannot be avoided. When using the extracted coordinates to locate each photo, we cannot guarantee full accuracy.

This study uses the Google Cloud Vision AI library to detect languages in each photo. We used the API's Optical Character Recognition (OCR) function to detect and extract text from signs in the photos. The extracted text strings were then passed through API's built-in language detection services, which attempts to classify the languages of each text segment. This workflow allowed rapid, large-scale annotation of hundreds of photos across multiple common languages, including English, Chinese, Japanese, Japanese and Vietnamese. The Google Cloud Vision AI library can handle multiple languages in one image without requiring separate models. Based on the proportion of each language appearing in the image, the library ranks the detected languages, with the top-ranked language representing the most dominant in the photo. When the classification is unclear, the model outputs "unknown", which is common for graffiti or complex street views. The output "N/A" indicates no languages are detected. The outputs are compared to human-labeled languages to evaluate accuracy and to identify common error patterns.

3. Results

3.1.     **Language recognition accuracy**

We collected a total of 1,449 photos from Honolulu Chinatown. After comparing the results from automatic language detection using Google Cloud Vision AI model and our manual checking, we found that 1,149 photos have been correctly labeled. This achieves an accuracy of 79%.

3.2.     **Mis-labeling**

   We identified five common categories of mistakes made by the Google Cloud Vision AI model during the language recognition process. In this section, we discuss each of these mis-labeling categories in detail and provide examples.

### 3.2.1. Distorted signs

These signs appear distorted either because of the photo's angle, the placement of the sign itself, or because they are not the main focus of the photos.

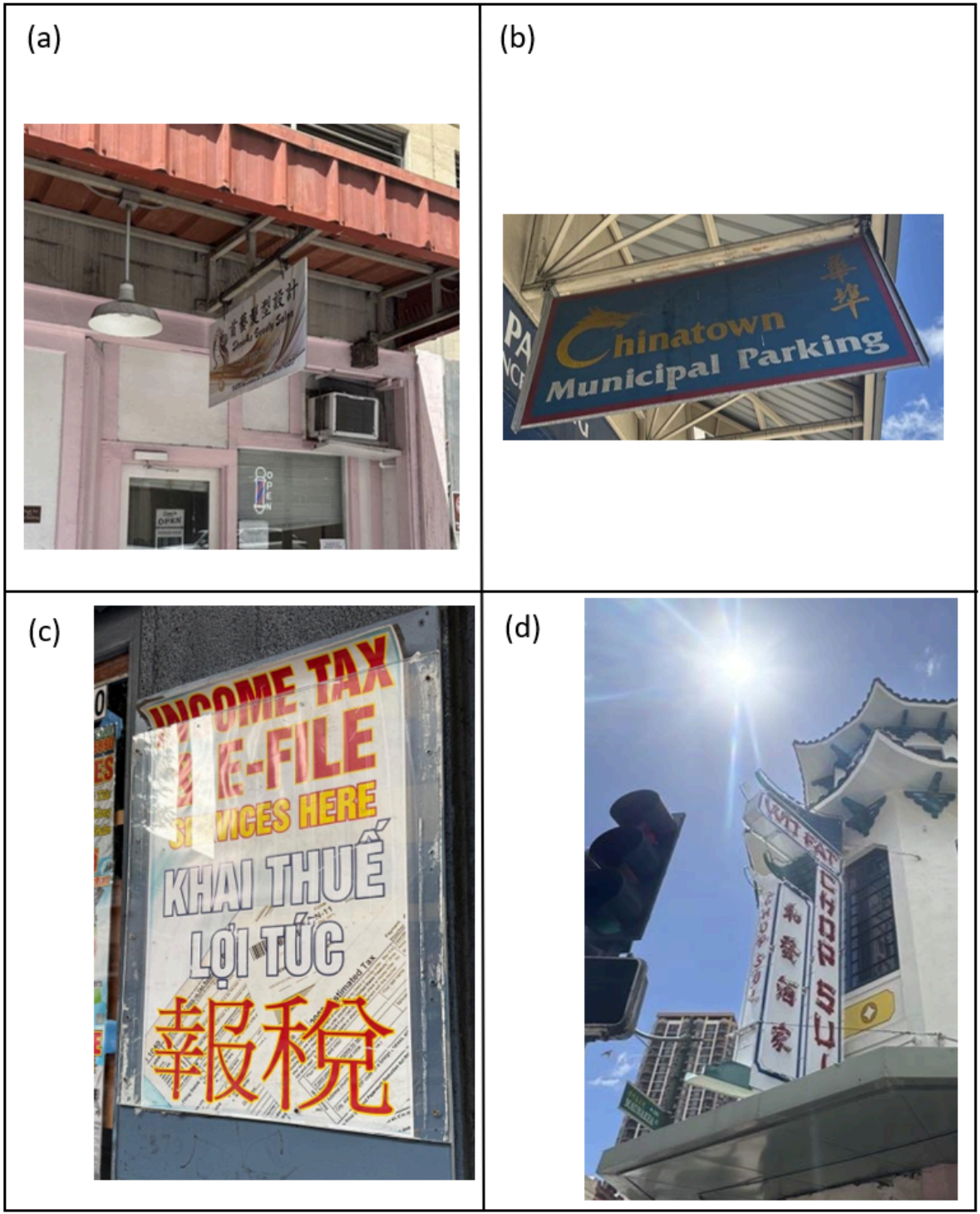

Figure 2. (a) A sign for a hair salon. (b) A hanging sign for "Chinatown Municipal Parking". (c) A poster for a tax service. (d) A restaurant sign.

Figure 2 (a-d) presents four examples of mis-labeled photographs within this category. All of these four pictures include Chinese characters indicating the store services. Figure 2(a) is a hair salon where the Chinese sign is the name of the salon. Because the sign hangs high above the sidewalk, the photo taken from a distance captures a steep angle. Figure 2(b), is a hanging sign for "Chinatown Municipal Parking" with two Chinese characters. The photo was taken from below, and the two Chinese characters have low color contrast, so the Google Cloud Vision model did not recognize Chinese in this photo and marked the second language as "N/A". Figure 2(c) is a poster for a tax service containing three languages: English, Vietnamese, and Chinese. The AI model correctly identified English and Vietnamese, despite reflections on the plastic cover. However, the model mis-labeled the clearly visible Chinese characters as "unknown". Figure 2(d) shows a restaurant sign hanging above the first-floor entrance door. Because the photo was taken at a non-orthogonal angle, the sign appears partially distorted.

It remains unclear why the AI model sometimes marks a language as "unknown" but sometimes marks as "N/A" when no more language is detected. Among all the photos, Google Cloud Vision AI model mis-labeled 19 photos due to this reason and all of these photos either did not detect Chinese characters (marked as "N/A") or failed to identify the characters are Chinese (marked as "unknown)".

### 3.2.2. Broken or ineligible parts

In Honolulu Chinatown, many businesses and service providers place signs on their glass windows or doors as stickers. Over time, these signs may become worn, faded, or dirty, which can make portions of the characters illegible. In other cases, photo lighting, reflections, or glare can cause strokes or parts of the stickers to appear incomplete in the photographs. Photographs that are mis-labeled by the AI model for these reasons are grouped in this category.

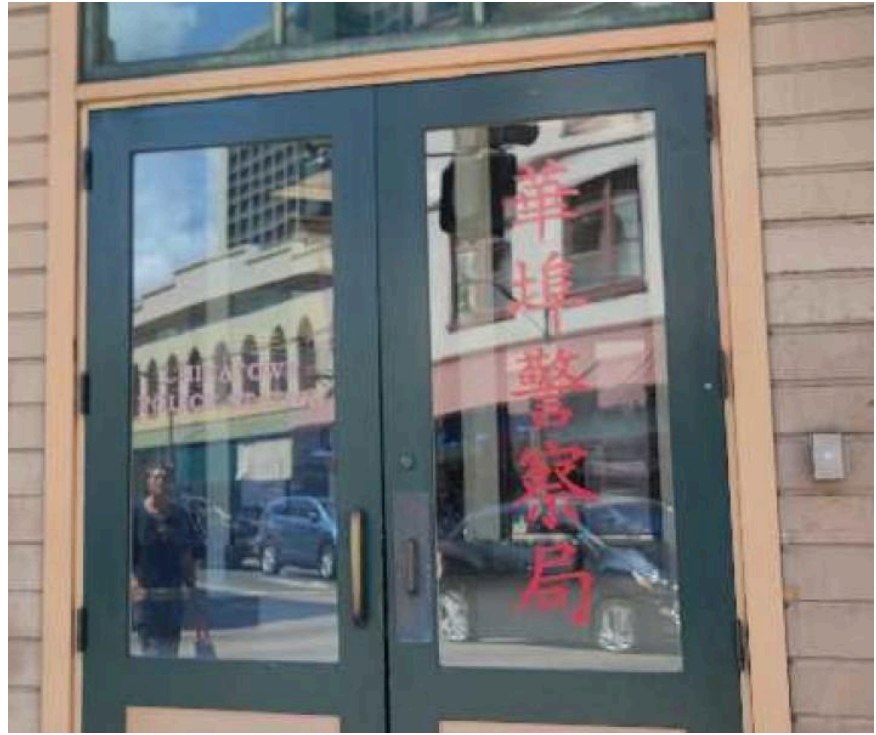
Figure 3. Honolulu Chinatown police station.

Figure 3 shows the glass door for Honolulu Chinatown police station. The signage appears in both English and Chinese, with the English name in a smaller font on the left door and the Chinese name in a larger font on the right door. Because the glass door is highly reflective, the signs are very difficult to read, especially the English text. As a result, Google Cloud Vision AI model labeled this picture as "N/A" for all the languages, indicating no languages were detected.

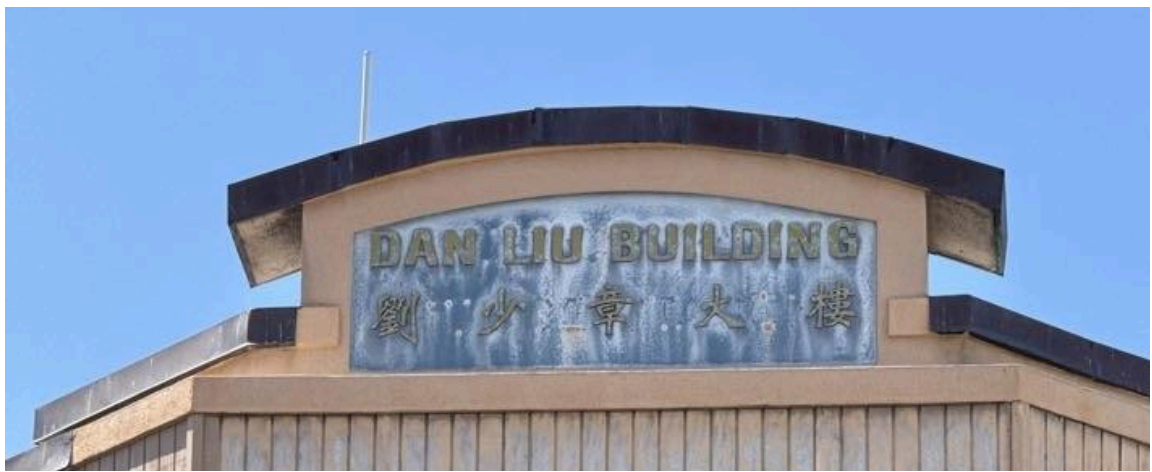
Figure 4. Dan Liu Building.

Figure 4 shows the English and Chinese names of a building. In this case, the Google Cloud Vision AI model labeled English as the only language in this picture. The original background color of the building was likely white; however, after long time exposure to weather, the surface appears dirty and has taken on a darker shade. This worn and darkened background likely made the sign harder to read and may have contributed to the model's mis-labeling.

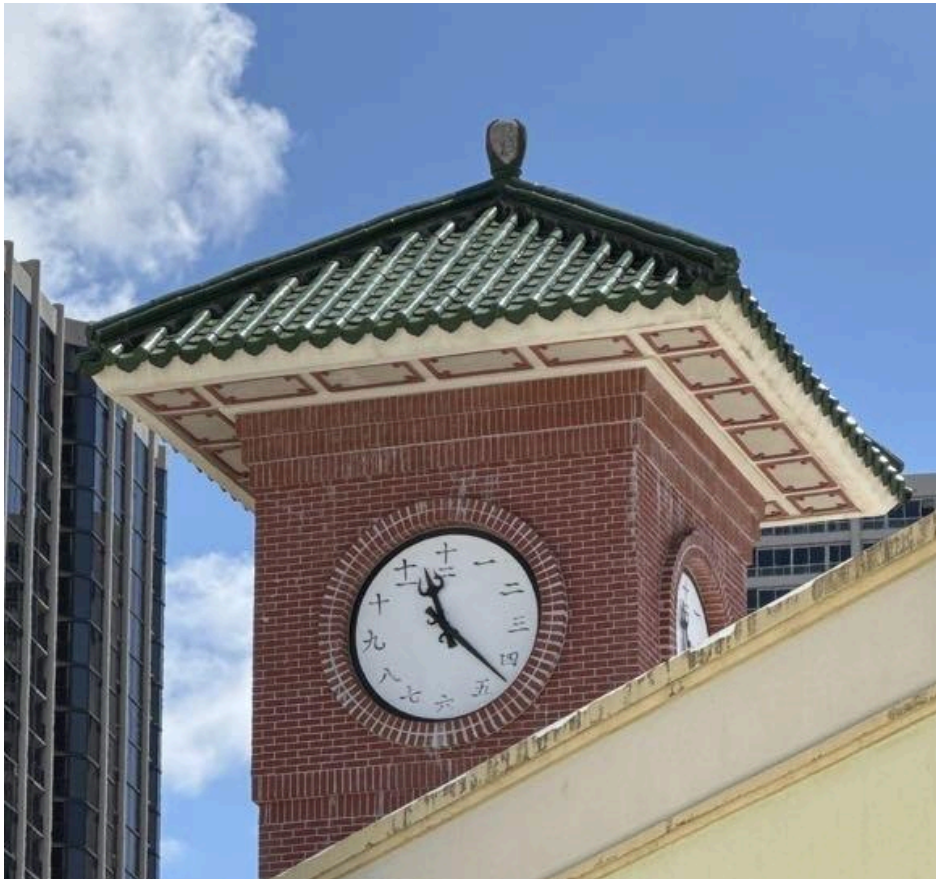

Figure 5. A clock with Chinese characters

Figure 5 shows the clock on top of a building located on N. Hotel Street. The AI model marked as "N/A", indicating that no languages were detected. However, the numbers from one to twelve on the clock face are written in Chinese characters. The failure to recognize the language may be due to parts of the characters being worn or obscured by the clock hands, or due to the model's difficulty in identifying Chinese when only isolated characters present.

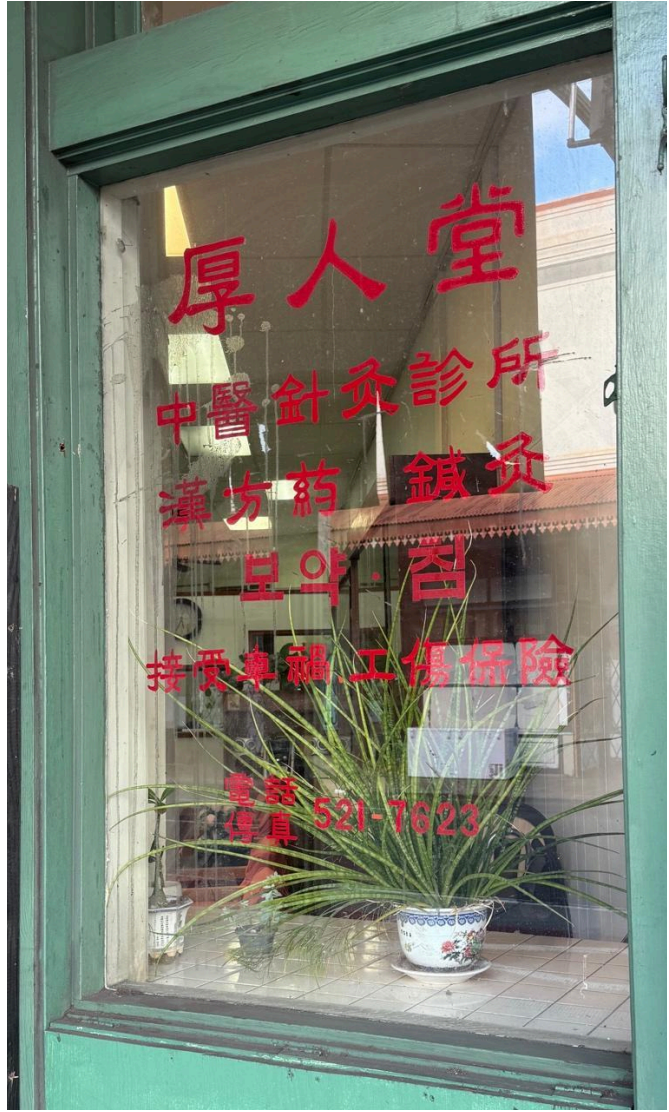

Figure 6. Window of a medical clinic.

Figure 6 shows signage for a medical clinic whose name and available service are displayed as stickers on the window. Google Cloud Vision AI model recognized English and Chinese in this picture, though no obvious English texts exist. Instead, three Korean characters appear on the window, which the model failed to recognize. Because these characters are applied as stickers and the window glass was reflective, the reflection likely contributed to the mis-label.

### 3.2.3. Graffiti

Graffiti is an important element of street culture. In Honolulu Chinatown, we observed a large amount of graffiti, including layers of spray paint over existing graffiti. As expected, we found Google Cloud Vision struggles to handle these photographs, particularly when the text appeared in irregular or stylized fonts, making accurate language detection hard even for human.

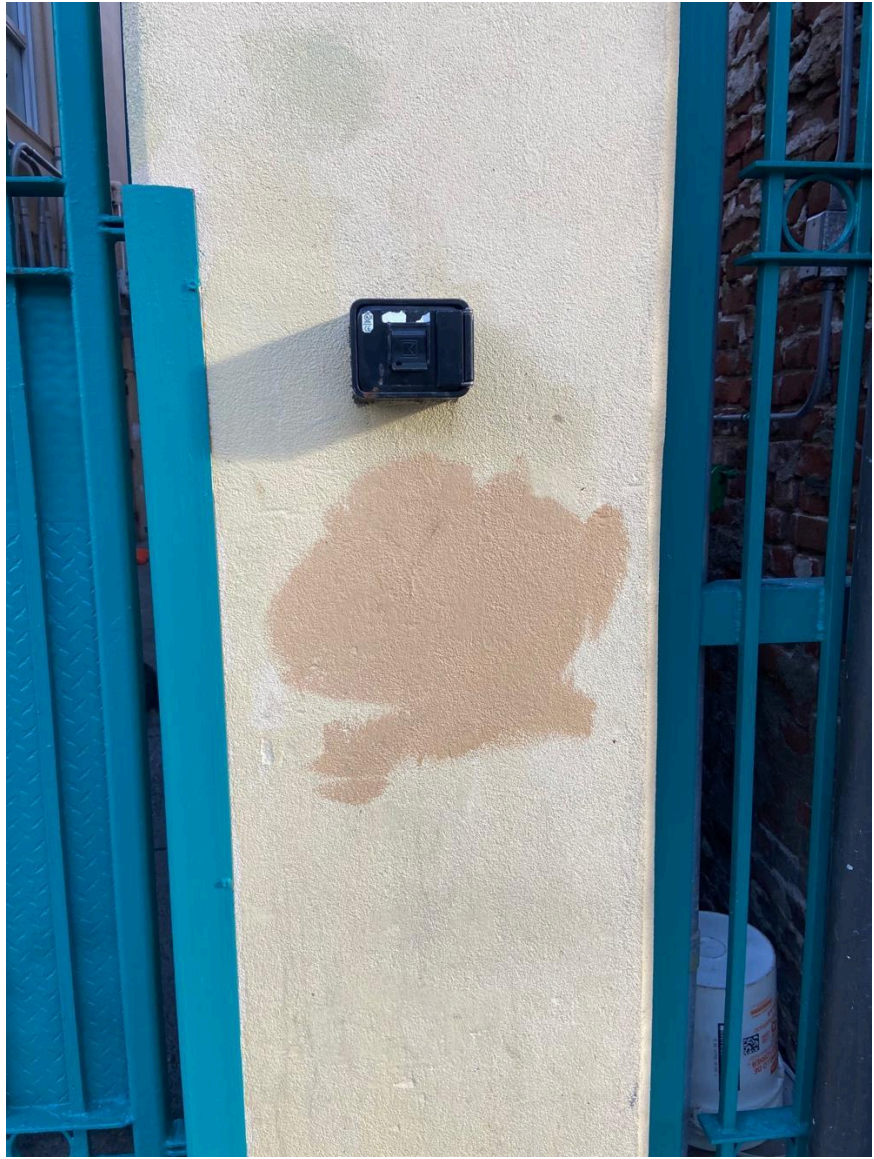

Figure 7. Wall with covered paints.

Figure 7 is an uncropped photo taken to document visible effort to cover something previously existing on the wall. During the language detection process, the AI model identified English as the only language in this photo. Although the object is not the focus of the picture, this result has been triggered by the up-side-down bucket at the bottom right corner of this picture.

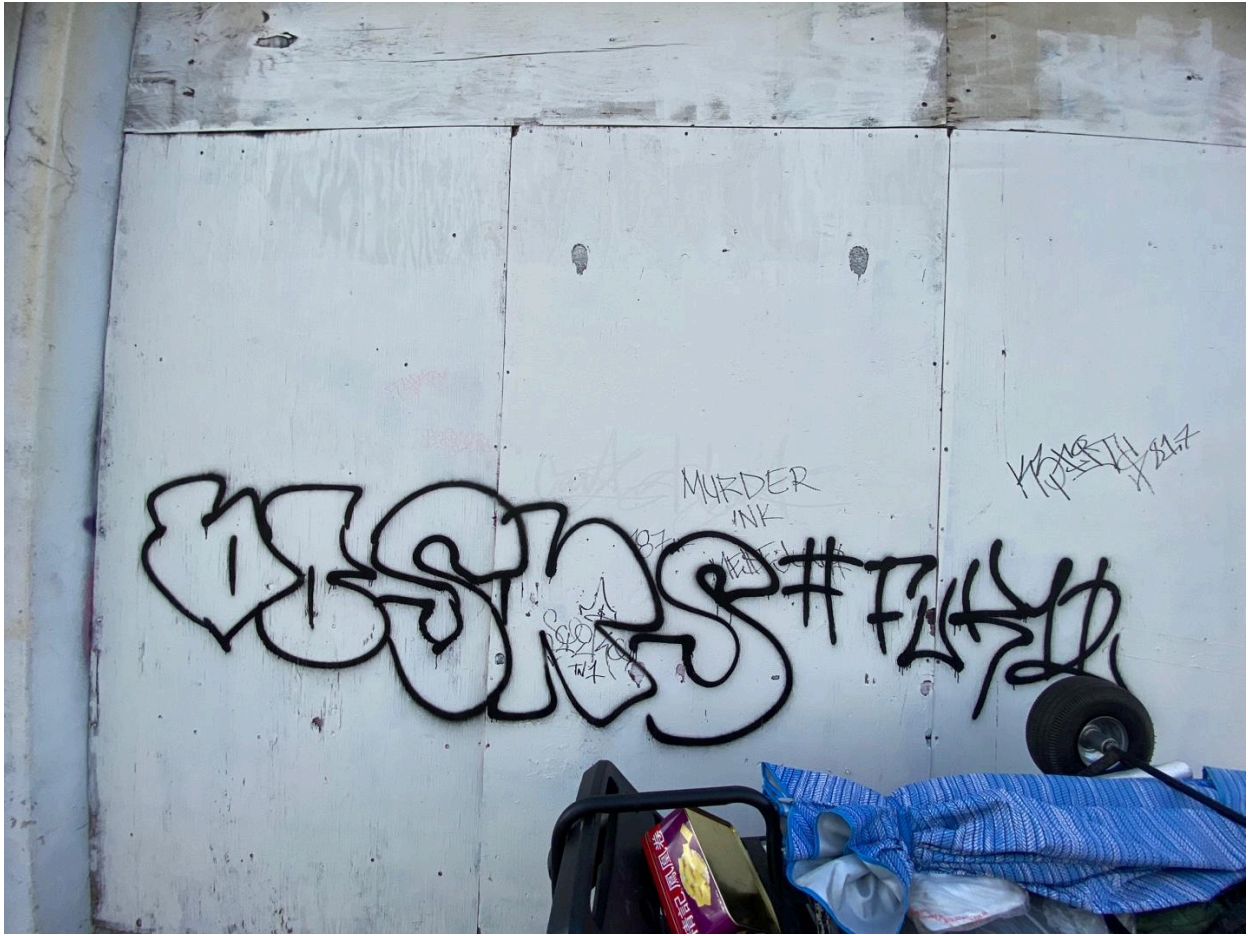

Figure 8. Graffiti.

Figure 8 is another uncropped photograph showing graffiti on Kekaulike Street. Google Cloud Vision AI model identified three languages found in this picture: English, Chinese, and unknown. This graffiti appears to be expressive and artistic, suggesting the creator(s) may have intended to convey a message through stylized lettering. Parts of the graffiti can be clear recognized as English letters, while other parts are unclear. The Chinese detected in this picture is not from the graffiti itself, but from a metal snack box at the bottom of this picture. During the model training process, our initial objective includes having the AI model to prioritize any recognized languages. In Figure 8, it marked English as the most dominant language and Chinese as the second. However, from a human perspective, we first recognize English and then an "unknown" language because the graffiti letters are hard to read. In our initial manual checking process, we did not notice the Chinese characters on the box because our focus was on the graffiti area. This example highlights a key consideration for AI-powered language detection and ranking systems, as the model may prioritize languages based on the length, regardless whether they are the focus of the picture or not. On the other hand, human reviewers need to carefully examine the entire image during manual verification to ensure the accurate checking, not just the main subject of interest.

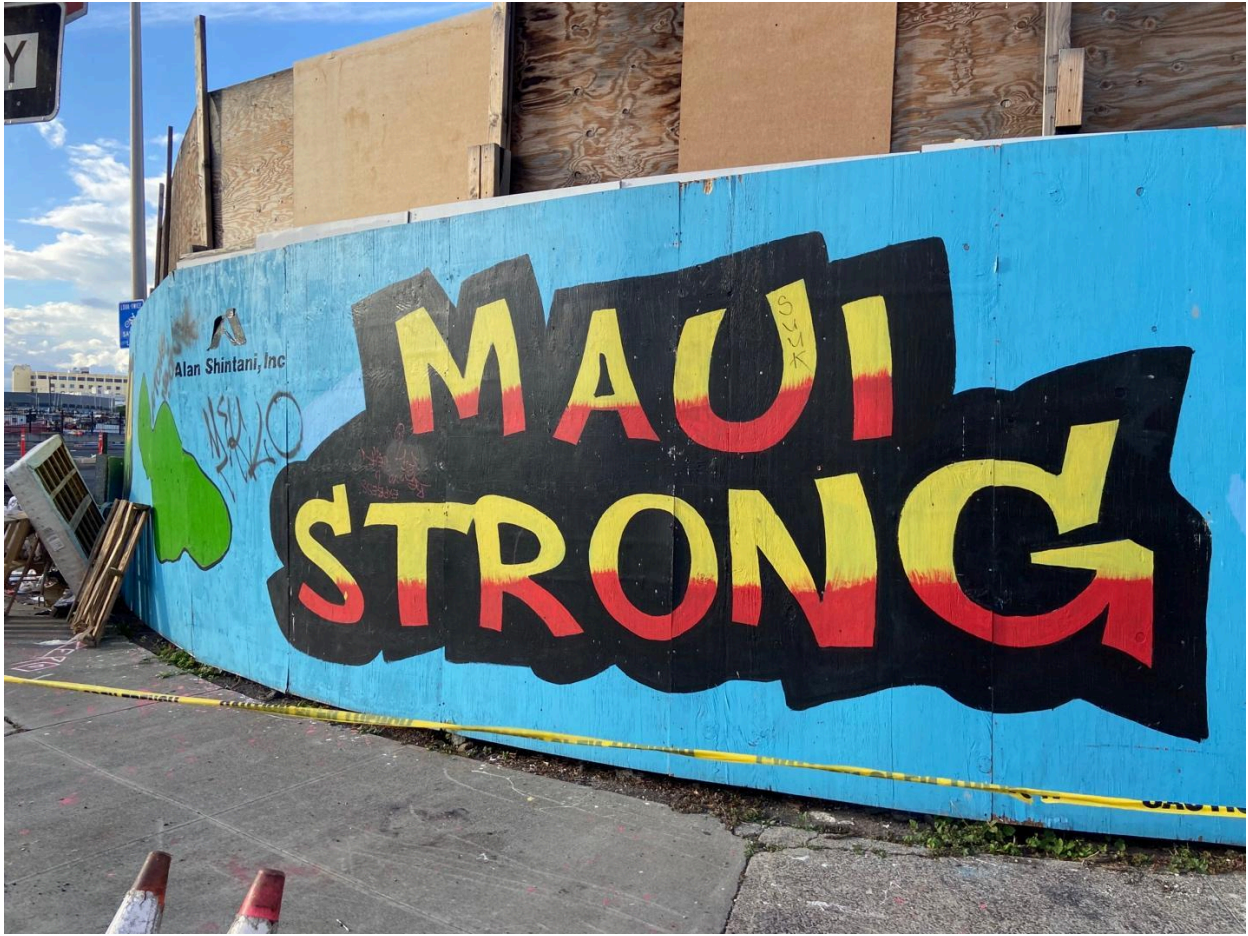

Figure 9. "Maui Strong" mural.

Figure 9 is a representative example of most photographs in this category. This photo captures an official "Maui Strong" mural, created to support for Maui communities affected by the 2023 Maui fires. However, on the blue background of the mural, someone has added a smaller ineligible piece of graffiti. The meaning and purpose of the smaller graffiti is unknown. We cannot recognize the language of this graffiti. The AI model labeled English as the only language present in this picture, but for our linguist landscape study, we must add "unknown" as an language for this photo to account for this small graffiti.

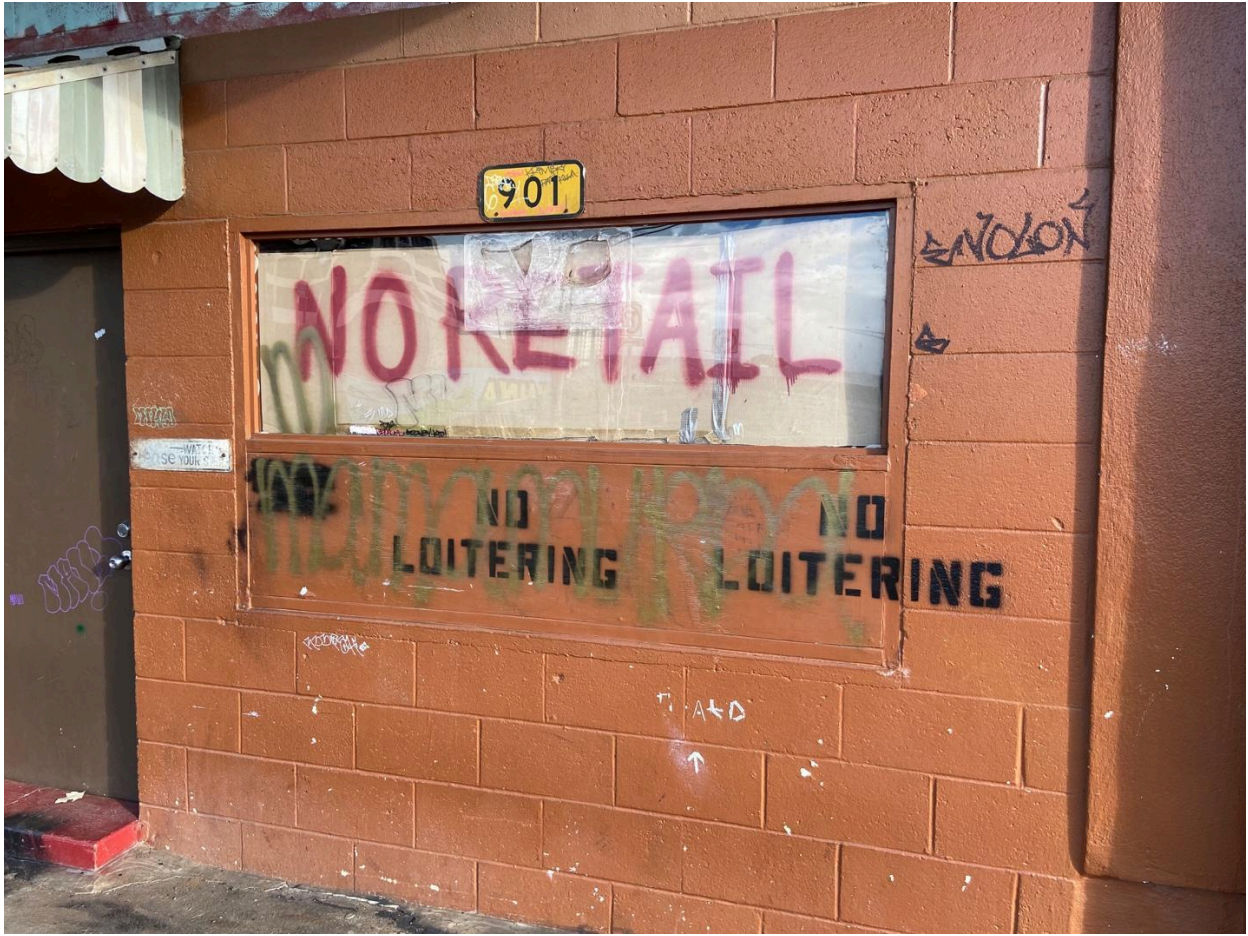

Figure 10. A wall with multiple signages.

Figure 10 illustrates a good example of the coexistence of official signage, unofficial signage, and graffiti. Two "NO LOITERING" notices were officially stenciled by the local authority on the building's external wall. Inside the window, the occupants of this housing unit have placed a large hand-written "NO RETAIL" sign, which is an unofficial but informative notice. On top of the "NO LOITERING" signs, spray-painted marks partially obscure the official signs. In addition, there is an ineligible graffiti mark to the right of this window. The graffiti seems to be composed of letters, but its meaning and purpose are unclear. Google Cloud Vision AI model results show English as the only language in this picture, recognizing the "NO LOITERING" and "NO RETAIL". Despite the "NO LOITERING" is partially covered by the graffiti and the "NO RETAIL" is affected by the window glass reflection, the AI model successfully recognized both signs, as a human would. However, for the purposes of our linguist landscape study, we manually changed "N/A" to "unknown" to indicate the presence of an ineligible language element in the picture, which was not successfully recognized by the AI model.

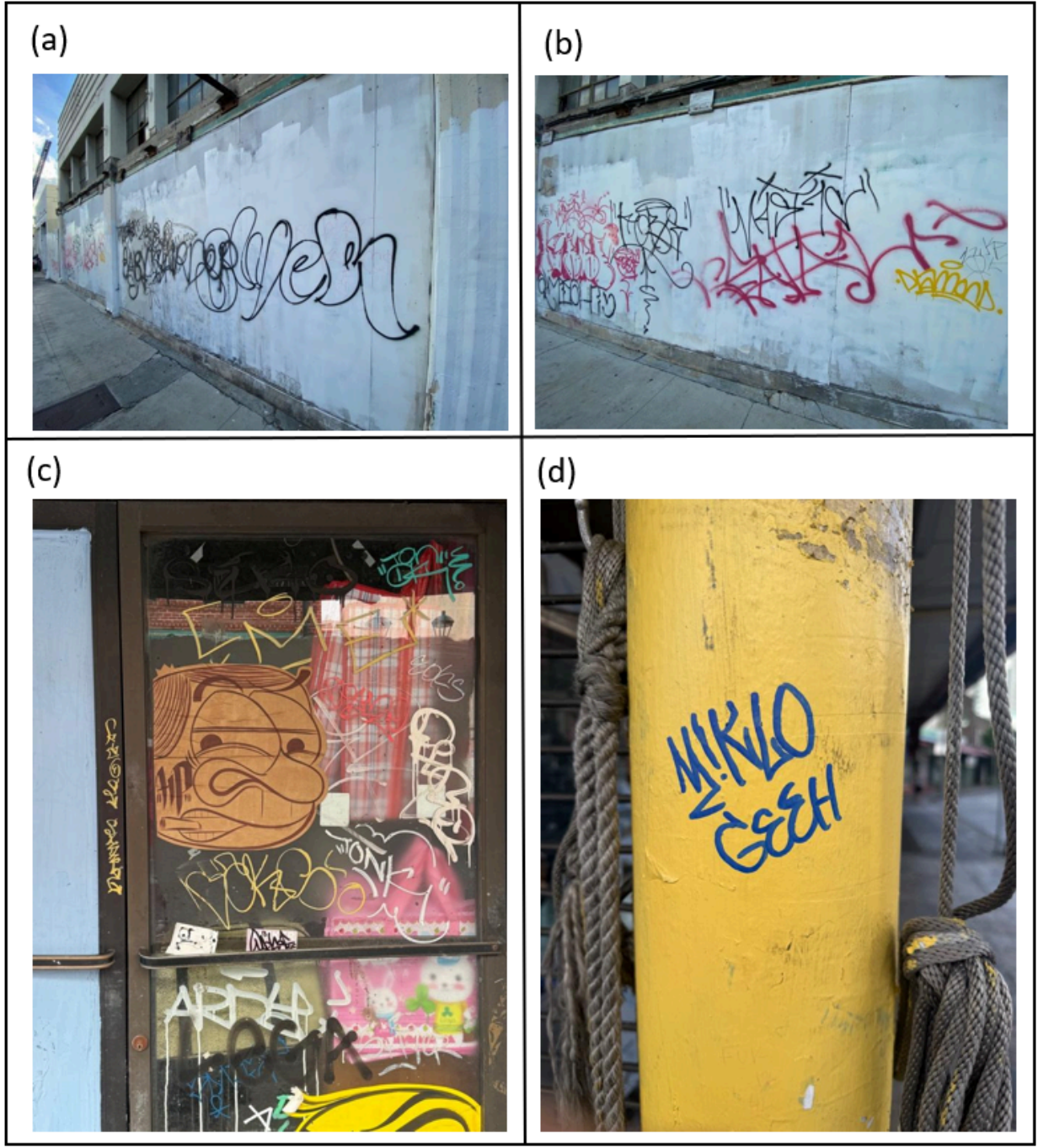

Figure 11 (a-d). Graffiti in Honolulu Chinatown.

Figure 11 (a-d) presents several examples of mis-labeled photographs caused by unclear graffiti or spray-painted markings. In the AI recognition process, Google Cloud Vision AI labeled English as the only language in all of these photos. During manual checking, however, we could only identify isolated letters, for example, letter "e" in Figure 11(a). In Figure 11(d), for instance, a hand-written phrase on a pillar appear to read "MIKLO GEEH", but several parts remain unclear and unverifiable. Because of the dominant patterns are in letter shapes, it is understandable that the AI model labeled those as English. However, from a human perspective, we reclassified those as "unknown", because the graffiti patterns did not convey any clear messages and consisted only of indistinct or ambiguous lettering.

*3.2.4. Japanese and Chinese mis-labeling*

Honolulu Chinatown is multi-cultural neighborhood with a mix of diverse Asian communities. In addition to Chinese, Japanese is also a very common language appearing on local signs. However, because some Japanese characters (Kanji) are visually identical to Chinese characters, we found it difficult for Google Cloud Vision AI model to correctly distinguish Japanese from Chinese, especially on signs with short phrases or limited context.

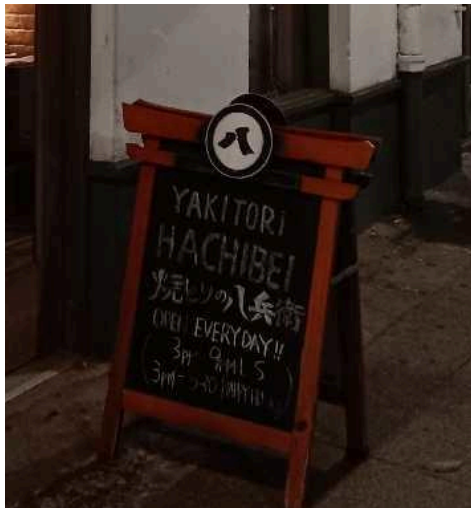

Figure 12. A hand-written sign outside a restaurant.

Figure 12 shows a hand-written sign placed outside a restaurant. The AI model labeled this picture as containing English and Chinese. However, instead of Chinese, the second language is actually Japanese. Although the characters identified by the AI model appear in both Chinese and Japanese, we are able to determine the correct language by noting the surrounding Japanese kana (hiragana and katakana), which provided clear contextual clues.

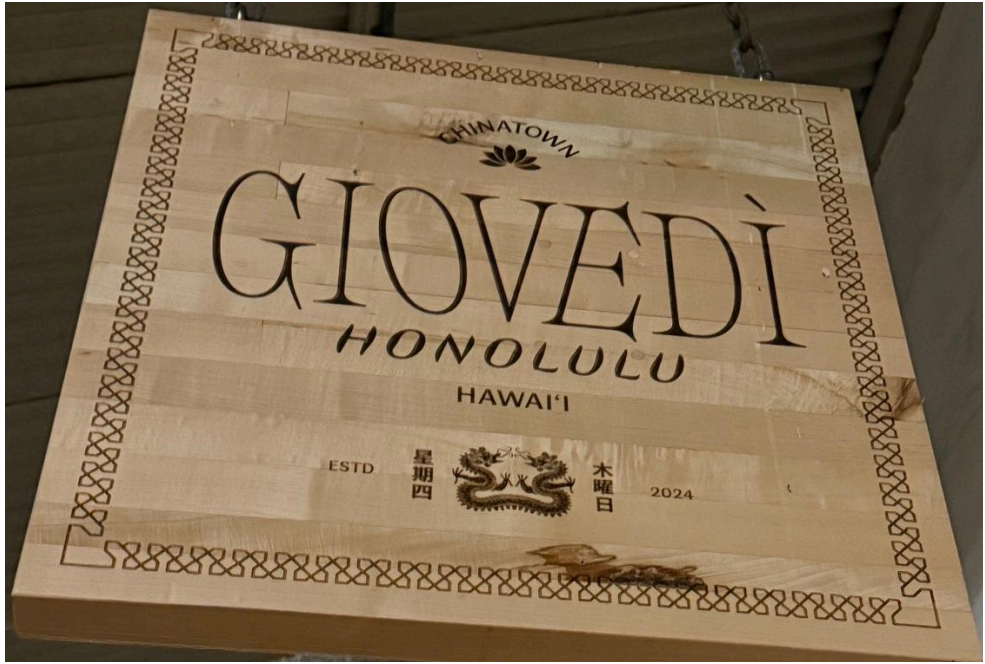

Figure 13. An overhead sign for a restaurant.

Figure 13 is an overhead hanging sign for a restaurant. Google Cloud Vision AI model labeled the languages in this picture are English and "unknown". However, upon closer examination of this photo, we found that, in addition to English, both Chinese and Japanese are present on the sign. Each side of the dragon at the bottom of the sign displays characters meaning "Thursday" in Chinese and Japanese. However, in this case, the AI model failed to detect both languages for unclear reasons.

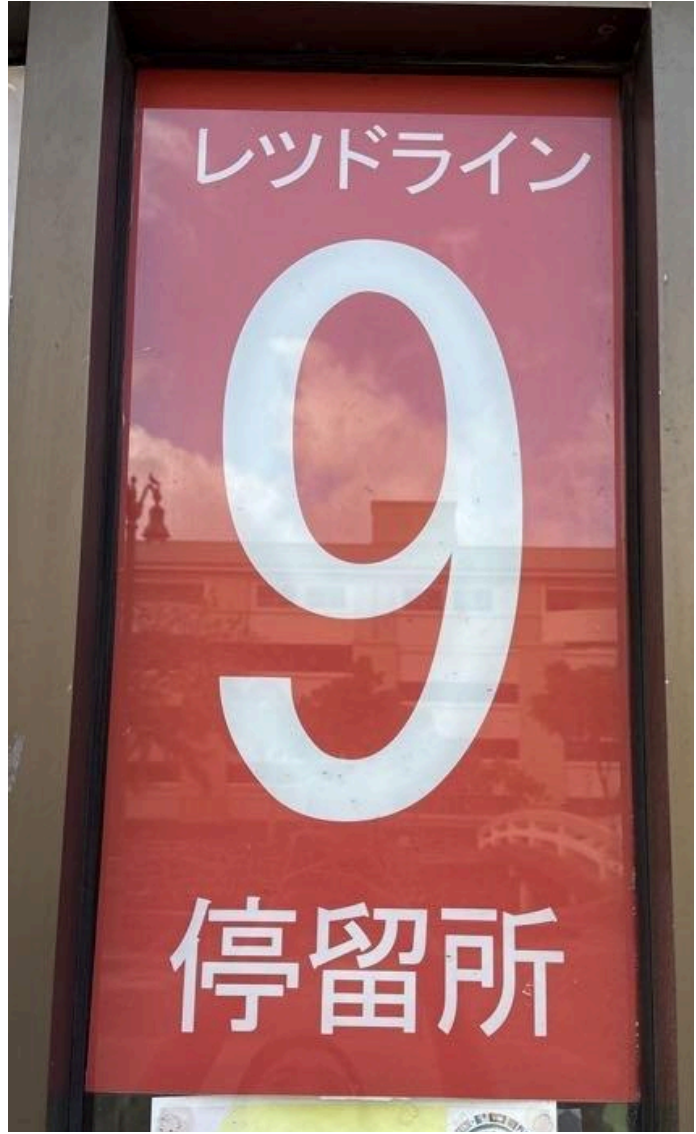

Figure 14. A sign posted by a restaurant.

Figure 14 shows a sign posted by a restaurant on River Street, which appears to designate a pickup or delivery spot for the restaurant. Google Cloud Vision AI model language detection results indicate that the sign contains both Japanese and Chinese. At the top, Japanese katakana characters were correctly identified by the model. At the bottom, however, there are three characters (kanji) meaning "stop (bus, tram, etc.), station, or stopping place" in Japanese. No Chinese language was presented in this photo. Similar to earlier examples, the AI model failed to "understand" the meaning or to distinguish Chinese and Japanese characters. For images or signs like this, accurately distinguishing the languages requires deeper knowledge of both writing systems and context, an ability appears to be beyond capabilities of current Google Cloud Vision AI mode.

3.2.5. *Hallucinated language*

During our manual checking process, we found that AI hallucination remains a major issue. In some photographs, Google Cloud Vision AI model detected languages that we could not verify during our manual checking process.

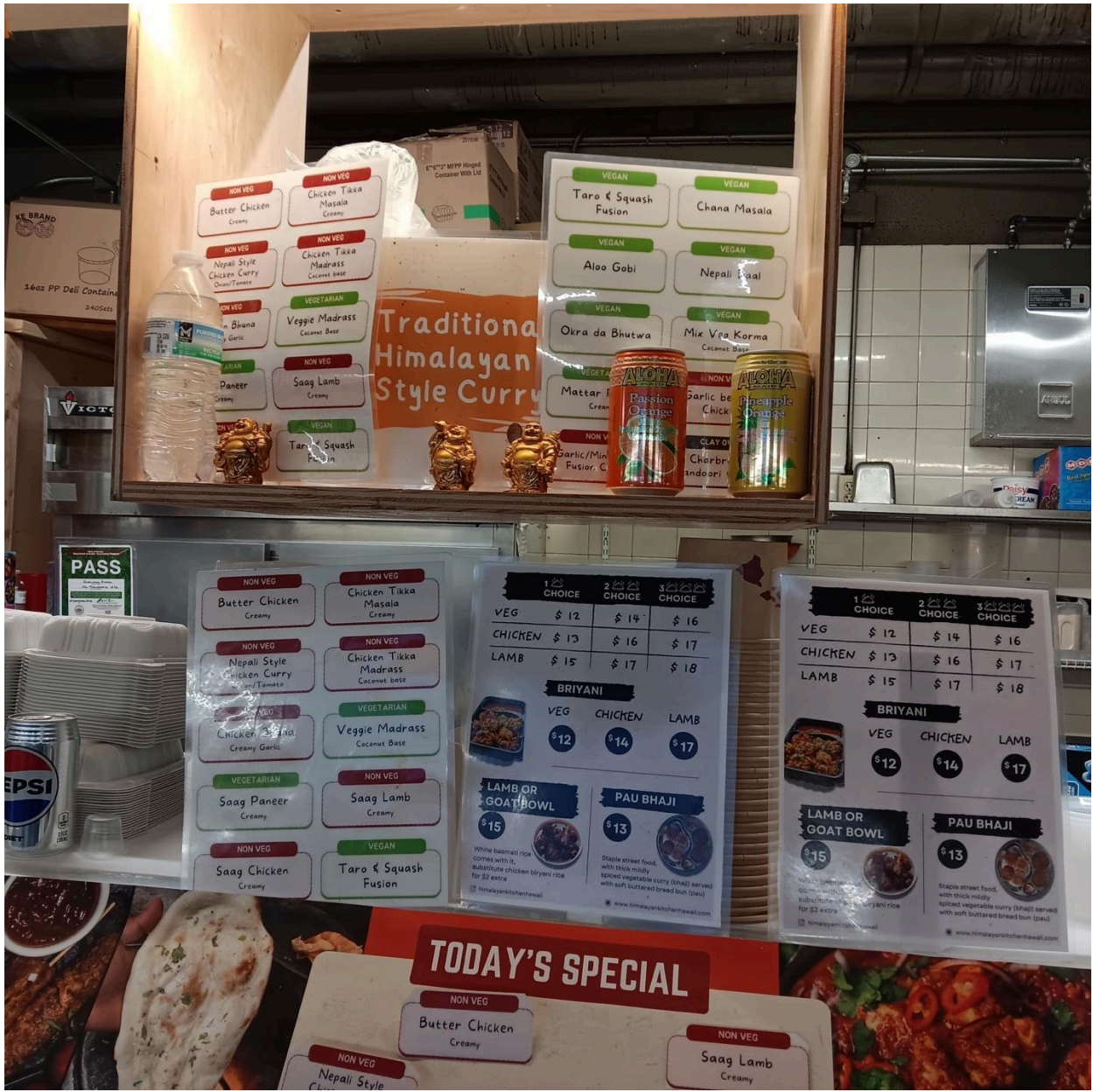

Figure 15. Menu of a restaurant.

Figure 15 shows a menu posted by a restaurant in Maunakea Marketplace. We took two photos of these menus that appears almost identical by our visual examination. During our manual review, we found English to be the only language present in both photos. However, Google Cloud Vision AI model identified Chinese in one of these two photos and Japanese in the other, despite there being no visible text in either language.

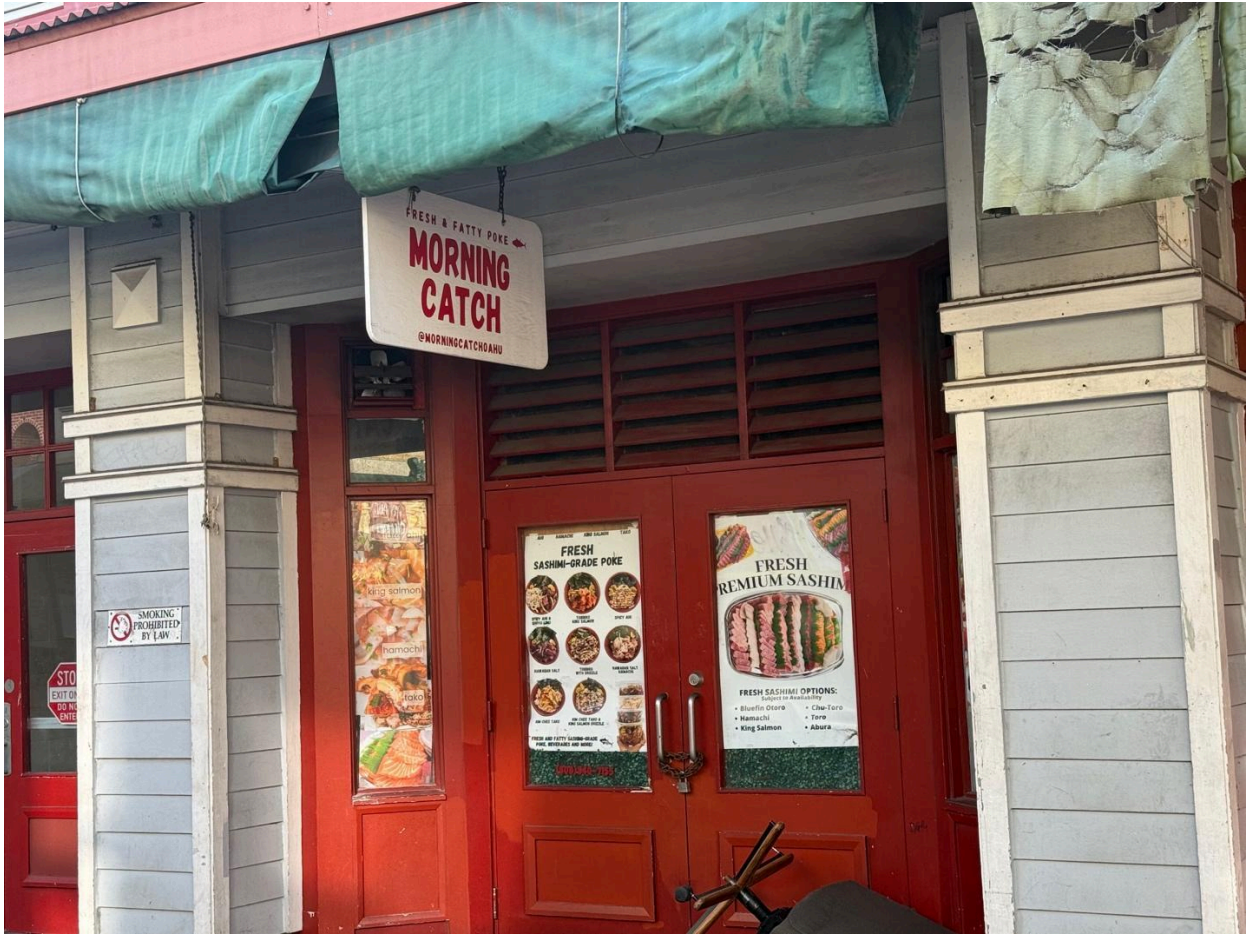

Figure 16. Menus and signages for a restaurant.

Figure 16 shows menus and signage for a restaurant. During our manual examination, we found English to be the only language in this photo. However, the AI model language detection results indicated an additional "unknown" language exists besides English. This appears to be another instance of AI hallucination, in which some unidentified elements of the image led the AI model to infer the presence of a non-existent language.

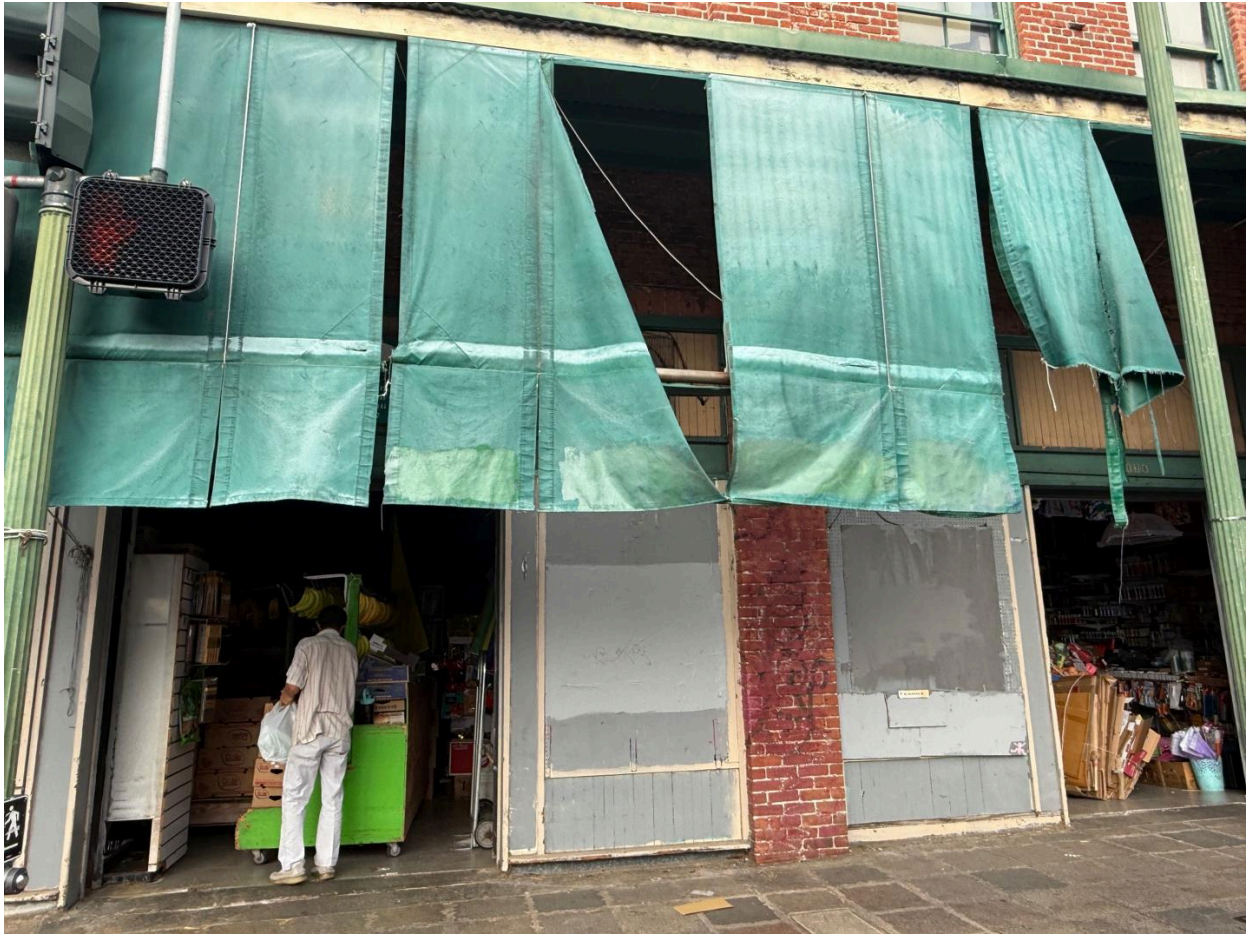

Figure 17. A street view photo.

Figure 17 is an example of street view photos we took to understand the general landscape of the neighborhood. There is no obvious visible language we identified from this photo. However, Google Cloud Vision AI model detected English in this picture. This occurred in multiple similar street-view photos. Although we do not identify any English text, the AI model frequently marked English as a language in such photos.

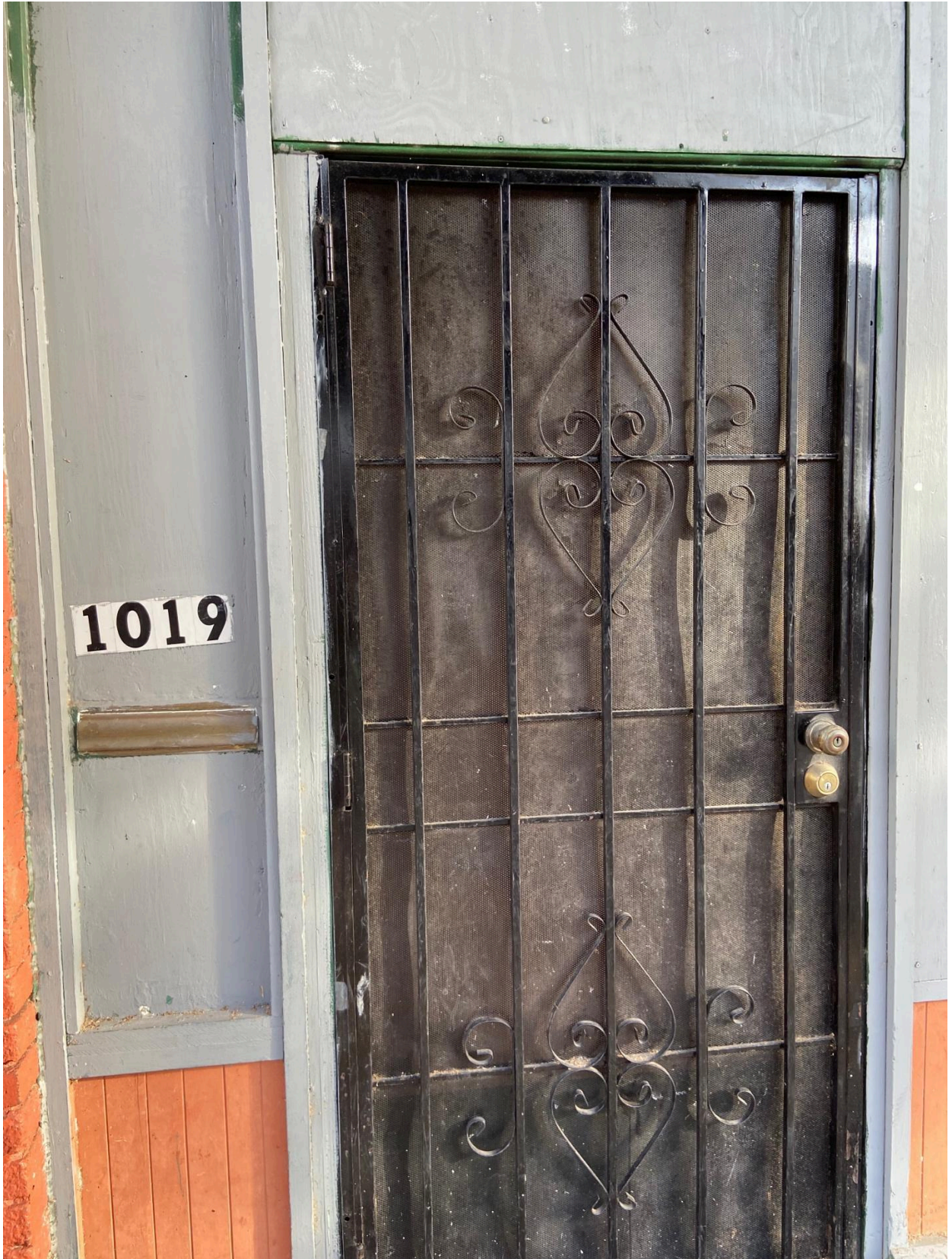

Figure 18. Entrance door to a residential unit.

Figure 18 shows the entrance door to a residential unit. The only signage visible in this photo is the number by the door indicating the unit number, which only contains number digits. However, the AI model detected English in this photo. It is not clear whether the model mistook numbers for English or the model coded numbers as part of the English language. It is also possible that some other element of the photo resembled English text, leading the model to label English as the dominant language in this photo.

## 4. Discussion and conclusion

Our findings show that the Google Cloud Vision AI model achieved an overall accuracy of 79% in recognizing signage languages across 1,449 photographs collected in Honolulu Chinatown. This level of accuracy demonstrates the potential feasibility of using AI tools for large-scale linguistic landscape analyses, yet it also highlights the current technological limitations when applied to multilingual and visually complex environment, especially with graffiti and irregular fonts. Mis-labeling was not random. Errors were frequently found from (1) visual distortion and non-orthogonal camera angles, (2) reflective or translucent materials such as glass, (3) textual ambiguity in graffiti or stylized fonts. Each of these factors affect the recognition process, causing subsequent language misclassification. The observed failure to distinguish between Chinese and Japanese characters further illustrates a systematic weakness of current models in handling language context. Finally, "AI hallucination" was detected, where languages that do not exist in the image were found. On the other hand, some texts in standardized fonts with clear backgrounds were mis-labeled.

From the perspective of LL methodology, the results highlight potential for automated language detection, which can significantly reduce human labor. Especially when the signages are in standardized fonts and photos are taken with no ambiguity. However, the types of errors revealed that AI outputs should NOT be treated as fully reliable. As we went through the manual check, we still found about 21% photos with mis-labeled languages. Without contextual and ethnographic interpretation, the model may misclassify languages. In addition, human and AI may have different focuses. Humans generally photograph with a central focus, such as a store sign, while treating the surrounding objects as non-important. However, the AI models process all areas of the image with an equal importance and may pick up textual or pattern-like elements at the edges, which are not the intended focus. Future studies along this direction can make several technical enhancements to improve the performance. First is model fine-tuning. This is specifically important to distinguish Chinese and Japanese. Our current model was not tuned with specific languages. Second, photos can be pre-processed to ensure the signages are the only focus of the photo. Other image enhancements, such as the change of brightness and contrast can be applied. Third, different AI models can also be applied for cross validation and performance evaluation. While this study only applies Google Cloud Vision AI model, other OCR-based AI language detection models can be applied for comparison and validation.

Honolulu Chinatown exemplifies a diverse urban environment where multiple Asian cultures intersect. This study demonstrates that even in such complex context, AI model can reveal language presence, but qualitative interpretation remains required for understanding sociolinguistic meanings. This research provides one empirical evaluation of AI-powered language detection in linguistic landscape analysis. Using Honolulu Chinatown as a case study, we demonstrate both the success and pitfalls of applying AI models to multilingual detection. The AI model exhibited multiple errors related to vision distortion, surface degradation, reflective surface, and other visual ambiguity. These findings indicate that fully automated LL analysis is not yet attainable. Instead, hybrid workflow and AI-assisted automation with human validation are more efficient and effective at the current stage. Finally, this work contributions to using computational AI and image vision technologies for sociolinguistic field research. This workflow increases efficiency for scalable and data-driven research agenda developments.

**Reference**

Amos, H. W. (2021). Chinatown by numbers: Defining an ethnic space by empirical linguistic landscape. Linguistic Landscape. An International Journal, 127–156. https://doi.org/10.1075/ll.2.2.02amo

Blommaert, J. (2013). Ethnography, superdiversity and linguistic landscapes: Chronicles of complexity. Multilingual Matters.

Coover, G. R. (2022). *Honolulu's Chinatown*. Rollston Press.

Jazul, M. E. M. A., & Bernardo, A. (2017). A look into Manila Chinatown's linguistic landscape: The role of language and language ideologies. Philippine Journal of Linguistics, 48(1), 75-98.

Leeman, J., & Modan, G. (2009). Commodified language in Chinatown: A contextualized approach to linguistic landscape 1. Journal of Sociolinguistics, 13(3), 332–362. https://doi.org/10.1111/j.1467-9841.2009.00409.x

Lou, Jackie. (2010). Chinese on the Side: The marginalization of Chinese in the linguistic and social landscapes of Chinatown in Washington, DC. In Shohamy, Elena, Ben-Rafael, Eliezer, & Barni, Monica (Eds.), Linguistic Landscape in the City. Channel View Publications.

Lou, J. (2012). Chinatown in Washington, DC: The bilingual landscape. World Englishes, 31(1), 34–47. https://doi.org/10.1111/j.1467-971X.2011.01740.x

Riley, N. E. (2024). Chinatown, Honolulu: Place, Race, and Empire. Columbia University Press.

Zhang, H., Tupas, R., & Norhaida, A. (2020). English-dominated Chinatown: A quantitative investigation of the linguistic landscape of Chinatown in Singapore. Journal of Asian Pacific Communication, 30(1–2), 273–289. https://doi.org/10.1075/japc.00052.zha